\pdfoutput=1
\PassOptionsToPackage{
pdfencoding=auto,
pdfnewwindow=true,
pdfusetitle=false,
pdfauthor={Noam Finkelstein},
pdftitle={Entropic Inequality Constraints from e-separation Relations},
bookmarks=true,
bookmarksnumbered=true,
bookmarksopen=true,
pdfpagemode=UseThumbs,
bookmarksopenlevel=1,
pdfpagelabels=false,
breaklinks=true,
backref=false,
colorlinks=true,
}{hyperref}
\PassOptionsToPackage{numbers,sort&compress,square,comma}{natbib}
\PassOptionsToPackage{sort&compress}{natbib}
\PassOptionsToPackage{usenames,dvipsnames,table}{xcolor}
\PassOptionsToPackage{skip=0pt}{caption}
\RequirePackage{etoolbox}
\patchcmd{\bibliographystyle}{#1}{apsrev4-2-wolfe}{}{}
\documentclass[accepted, mathfont=cm]{uai2021} %

\usepackage[american]{babel}

\usepackage{natbib} %
\usepackage{mathtools} %
\usepackage{booktabs} %

\AtBeginDocument{%
    \newwrite\bibnotes
    \def\bibnotesext{Notes.bib}
    \immediate\openout\bibnotes=\jobname\bibnotesext
    \immediate\write\bibnotes{@CONTROL{REVTEX42Control}}
    \immediate\write\bibnotes{@CONTROL{%
    apsrev42Control,editor="0",pages="0",title="0",year="1"}}
     \if@filesw
     \immediate\write\@auxout{\string\citation{apsrev42Control}}%
    \fi
}
\usepackage{amsmath,amsfonts,amssymb,amsthm,mathtools,bm,graphicx,verbatim,mathrsfs,units}
\graphicspath{{./}}

\usepackage{pgfplots}
\pgfplotsset{compat=newest}
\usepackage{tikzscale}
\usetikzlibrary{shapes.geometric}

\newcommand{\nocontentsline}[3]{}
\newcommand{\tocless}[2]{\bgroup\let\addcontentsline=\nocontentsline#1{#2}\egroup}

\DeclarePairedDelimiter\parens{\lparen}{\rparen}

\DeclarePairedDelimiter\bracks{\lbrack}{\rbrack}
\DeclarePairedDelimiterX\parensdo[2]{\lparen}{\rparen}{#1 \delimsize\vert #2}

\usepackage{bbold}

\usepackage[normalem]{ulem}
\newcommand{\st}[1]{\ifmmode\text{\sout{\ensuremath{#1}}}\else\sout{#1}\fi}

\usepackage{subcaption}
\usepackage[labelformat=simple]{subcaption}

\usepackage{thmtools,thm-restate}
\usepackage{color}
\usepackage{colordvi}

\usepackage[usenames,dvipsnames,table]{xcolor}
\usepackage{placeins} %

\definecolor{blue}{rgb}{0,0,1}
\definecolor{grey}{rgb}{0.6,0.6,0.6}

\definecolor{myurlcolor}{rgb}{0,0,0.7}
\definecolor{myrefcolor}{rgb}{0.8,0,0}
\definecolor{purple}{RGB}{128,0,128}
\definecolor{ultramarine}{RGB}{63, 0, 255}
\definecolor{medblue}{RGB}{0, 0, 100}
\definecolor{googleblue}{RGB}{34, 0, 204}
\definecolor{panblue}{RGB}{0,24,150}
\definecolor{carmine}{RGB}{150, 0, 24}
\definecolor{gray}{RGB}{150, 150, 150}

\usepackage{hyperref}
\hypersetup{
unicode=true,
bookmarksnumbered=false,
bookmarksopen=false,
breaklinks=false,
colorlinks=true,
linkcolor=carmine,
citecolor=googleblue,
urlcolor=panblue,
anchorcolor=OliveGreen}

\usepackage{paralist}

\newtheorem{thm}{Theorem}
\newtheorem{definition}[thm]{Definition}
\newtheorem{remark}[thm]{Remark}
\newtheorem{prop}[thm]{Proposition}
\newtheorem{lemma}[thm]{Lemma}
\newtheorem{cor}{Corollary}[thm]

\newtheoremstyle{defblock}{0.7\topsep}{0pt}{}{}{\color{medblue}\bfseries}{: }{0pt plus 1pt minus 1pt}{\thmname{\bfseries{#1}}\thmnumber{\bfseries{#2}}\thmnote{#3}}
\theoremstyle{defblock}

\theoremstyle{remark}
\newtheorem{example}{Example}

\usepackage{microtype}
\microtypecontext{spacing=nonfrench}
\microtypesetup{
expansion={true,nocompatibility},
protrusion={true,nocompatibility},
activate={true,nocompatibility},
tracking=true,
kerning=true,
spacing={true}
}

\usepackage[all=normal,floats=tight,mathspacing=tight,wordspacing=tight,paragraphs=normal,tracking=tight,charwidths=tight,mathdisplays=normal,sections=normal,margins=normal]{savetrees}
\usepackage[nodisplayskipstretch]{setspace}

\everypar=\expandafter{\the\everypar\loosness=-1 }
\usepackage[style=american,autopunct=true]{csquotes}
\usepackage[capitalise,nameinlink]{cleveref}
\Crefname{eqs}{Eqs.}{Eqs.}
\creflabelformat{eqs}{(#2#1#3)}
\Crefname{thm}{Thm.}{Thms.}
\crefrangelabelformat{eqs}{(#3#1#4-#5#2#6)}
\Crefmultiformat{eqs}{Eqs.~(#2#1#3}{,#2#1#3)}{,#2#1#3}{,#2#1#3)}
\usepackage{multirow}
\usepackage{array}
\newcolumntype{M}{>{\(}c<{\)}} %

\newcommand{\A}{{\boldsymbol A}}
\renewcommand{\a}{{\boldsymbol a}}
\newcommand{\B}{{\boldsymbol B}}
\renewcommand{\b}{{\boldsymbol b}}
\let\C\undefined
\newcommand{\C}{{\boldsymbol C}}
\renewcommand{\c}{{\boldsymbol c}}
\newcommand{\D}{{\boldsymbol D}}
\renewcommand{\d}{{\boldsymbol d}}
\newcommand{\X}{{\boldsymbol X}}
\newcommand{\x}{{\boldsymbol x}}
\newcommand{\Y}{{\boldsymbol Y}}

\newcommand{\V}{{\boldsymbol V}}
\newcommand{\Ub}{{\boldsymbol U}}

\newcommand{\Z}{{\boldsymbol Z}}
\newcommand{\z}{{\boldsymbol z}}
\newcommand{\DS}{{\boldsymbol D}^\#}
\newcommand{\ds}{{\boldsymbol d}^\#}
\let\G\undefined
\newcommand{\G}{\mathcal{G}}
\newcommand{\GS}{\mathcal{G}^{\#}}
\newcommand{\esep}{${(\A\perp_e \B\mid\C \text{ upon } \lnot \D)}$}

\usepackage{xspace}

\newcommand{\MME}[0]{\ensuremath{\operatorname{\textsf{\upshape MME}}}\xspace}

\newcommand{\ACE}[0]{\ensuremath{\operatorname{\textsf{\upshape ACE}}}\xspace}

\usepackage{mathabx}

\title{
  Entropic Inequality Constraints from \(\boldsymbol{e}\)-separation Relations in\\
  Directed Acyclic Graphs with Hidden Variables
}

\author[1]{\href{mailto:<noam@jhu.edu>}{Noam Finkelstein}}
\author[2, 3]{Beata Zjawin}
\author[2]{Elie Wolfe}
\author[1]{Ilya Shpitser}
\author[2]{Robert W. Spekkens}
\affil[1]{
  Johns Hopkins University, Department of Computer Science\\
  3400 N Charles St, Baltimore, MD USA, 21218
}
\affil[2]{
  Perimeter Institute for Theoretical Physics\\
  31 Caroline St. N, Waterloo, Ontario, Canada, N2L 2Y5
}
\affil[3]{
  International Centre for Theory of Quantum Technologies, University of\\
  Gda\'nsk, 80-308 Gda\'nsk, Poland
}

\begin{document}
\maketitle

\begin{abstract}
  Directed acyclic graphs (DAGs) with hidden variables are often used to
  characterize causal relations between variables in a system. When some
  variables are unobserved, DAGs imply a notoriously complicated set of
  constraints on the distribution of observed variables. In this work, we
  present entropic inequality constraints that are implied by \(e\)-separation
  relations in hidden variable DAGs with discrete observed variables. The constraints can intuitively be
  understood to follow from the fact that the capacity of variables along a
  causal pathway to convey information is restricted by their entropy; e.g. at
  the extreme case, a variable with entropy $0$ can convey no information. We
  show how these constraints can be used to learn about the true causal model
  from an observed data distribution. In addition, we propose a measure
  of causal influence called the minimal mediary entropy, and demonstrate
  that it can augment traditional measures such as the average
  causal effect.
\end{abstract}

\section{Introduction}

A causal model of a system of random variables can be represented as a directed
acyclic graph (DAG), where an edge from a node $X$ to a node $Y$ can be taken to
mean that the random variable $X$ is a direct cause of the random variable $Y$.
Such causal models can be used to algorithmically deduce highly non-obvious
properties of the system. For example, it is possible to deduce that the
probability distribution of observed variables in the system, called the
\textit{observed data distribution}, must satisfy certain \textit{constraints}.

When some variables in the system are unobserved, the constraints implied by the
causal model are not well understood, and, for computational reasons, cannot be
feasibly enumerated in full for arbitrary graphs. As a result, a number of
methods have been developed for quickly providing a subset of these constraints
\citep{wolfe2016inflation, KangTian2006, Poderini2019exclusivity}. In this work,
we contribute to this literature by describing entropic inequality constraints
that hold whenever an \(e\)-separation relationship \citep{Evans2012instrumental,
  Pienaar2017} is present in the graph.

The idea underlying these inequality constraints is that mutual information
between two variables in a graphical model must be explained by variability of variables (termed bottleneck variables) that are between them along some path.
Such paths need not be directed; a bottleneck variable may constitute the base of a fork structure or the mediary variable in a chain structure along the path.
Each such path has a limited capacity for carrying information, which can
be quantified in terms of the entropies of the bottleneck variables on that path. At the
extreme case, if there is a bottleneck variable along a %
path with zero entropy,
then subsequent variables on that path cannot learn about prior variables through the path, because the bottleneck variable will hold a fixed value%
regardless of the values taken other any other variables, observed or unobserved.
We will quantitatively relate the amount of information that can flow through a path to the entropies of its bottleneck variables below.

Constraints on the observed data distribution implied by a causal model have
primarily been used to determine whether the observed data is
\textit{compatible} with a causal model, and to learn the true causal model
directly from the observed data. Existing algorithms for learning causal models
rely primarily on equality constraints. We suggest that incorporating our
proposed inequality constraints, which can easily be read off a graphical
model, can meaningfully improve these methods. In addition, we show how the
entropy of latent variables can be linked to properties of the observed data
distribution, yielding bounds on latent variable entropies or constraints on the
observed data distribution.

We also demonstrate that our constraints can be used to bound an intuitive
measure of the \textit{strength} of a causal relationship between two variables,
called the \emph{Minimum Mediary Entropy} (\MME). We show that the standard
measure, called the \emph{Average Causal Effect} (\ACE), is not well suited
to capturing the causal influence strength of a non-binary treatment on outcome,
and can be misleading in some settings. For example, the \ACE can be $0$ even
when treatment changes outcome for every subject in the population. The \MME
overcomes both of these issues, and can serve as an informative complement to
the \ACE.

The remainder of the paper is organized as follows. In Section~\ref{sec:prelim},
we discuss relevant material in causal inference and information theory. We present
our constraints in Section~\ref{sec:constraints}, and several applications of
the constraints in Section~\ref{sec:applications}. Finally, a discussion of
related work and directions for future study can be found in
Section~\ref{sec:related-work} and Section~\ref{sec:conclusion} respectively.

\section{Preliminaries}
\label{sec:prelim}

\subsection{Causal Inference Background}

We begin by introducing key ideas from the literature on graphical causal
models. Suppose we are interested in a system of related phenomena, each of
which can be represented by a random variable. We denote observed variables in
the system as $\Y$, unobserved variables as $\Ub$, and the full set of variables
as $\V \equiv \Y \cup \Ub$.

We let $\mathcal G$ denote a DAG representing the system of interest. Each node
in $\mathcal G$ corresponds to a variable in $\V$. The direct causes of each
random variable $V$ are defined to be its parents in $\mathcal G$, denoted
$pa_{\mathcal G}(V)$. We adopt a nonparametric structural equations view of the
DAG~\citep{pearl2009causality, Richardson2013SingleWI}, under which the value of each variable $V$ is
a function of its direct causes and exogenous noise, denoted $\epsilon_V$. The
set of these structural equations is denoted $\mathcal F \equiv
\{f_V(pa_{\mathcal G}(V), \epsilon_V) \mid V \in {\V}\}$.  In most causal
analyses, the exact form of these functions is unknown. Nevertheless, if the
structure of causal dependencies in a system is known to be summarized by a
graph $\mathcal G$, or, equivalently, to be described by some set of functions
$\mathcal F$, then the distribution ${P({\V})}$ is know to factorize as
\begin{align}
  \label{eq:SEM}
  P({\V}) = \prod_{V \in {\V}} P(V \mid pa_{\mathcal G}(V)).
\end{align}
Equation~\eqref{eq:SEM} is the fundamental constraint that $\mathcal G$ places on the
distribution ${P({\V})}$ -- if the equality holds, then the distribution is in
the model; otherwise it is not. When all variables are observed, each term in
the factorization is identifiable from observed data, and the constraint may
easily be checked. When not all variables are observed, there is no known
polynomial-time algorithm for expressing the constraints that the factorization
of the full joint distribution places on the observed data distribution. In
theory, necessary and sufficient conditions for the observed data distribution
to be in the model can be obtained through the use of quantifier elimination
algorithms~\citep{geiger1999quantifier}, but these have doubly exponential runtime and are prohibitively slow
in practice.

We now review \(d\)-separation and \(e\)-separation, which are properties of the
graph $\mathcal G$ that imply certain properties of distribution ${P(\V)}$. We
first introduce the notion of open and closed paths in conditional
distributions. Triples in the graph of the form $A \rightarrow C \rightarrow B$
and $A \leftarrow C \rightarrow B$ are said to be open if we do not condition on
$C$, and closed if we do condition on $C$. Triples of the form $A \rightarrow C
\leftarrow B$, in which $C$ is called a collider, are closed if we do not
condition on $C$ or any of its descendants, and open if we do. A path is said to
be open under a conditioning set $\C$ if all contiguous triples along that path
are open under that conditioning set.

\begin{definition}[\(d\)-separation]
  Let $\A$, $\B$ and $\C$ be sets of variables in a DAG. $\A$ and
  $\B$ are said to be \(d\)-separated by $\C$ if all paths between $\A$ and
  $\B$ are closed after conditioning on $\C$. This is denoted
  ${(\A \perp_d \B \mid \C)}$.
\end{definition}

It is a well-known consequence of Equation~(\ref{eq:SEM}) that any \(d\)-separation
relation ${(\A \perp_d \B \mid \C)}$ in $\mathcal{G}$ implies the corresponding
conditional independence relation $\A \perp \B \mid \C$ in the distribution ${P(\V)}$.
Conditional independence constraints of this form are about sub-populations in
which the variables in $\C$ take the same value for all subjects. We can only
evaluate whether these constraints hold when all variables in $\C$ are observed;
otherwise there is no way to identify the relevant sub-populations. For that
reason, it is impossible to determine whether conditional independences implied
by $\mathcal G$ hold if they have hidden variables in their conditioning sets,
leading to the need for other mechanisms to test implications of these
independencies.

To describe \(e\)-separation, we first introduce the idea that a node can
be \textit{deleted} from a graph by removing the node and all of its incoming
and outgoing edges. \(e\)-separation can then be defined as follows.

\begin{definition}[\(e\)-separation]
  Let $\A$, $\B$, $\C$ and $\D$ be sets of variables in a DAG. $\A$ and $\B$ are
  said to be \(e\)-separated by $\C$ after deletion of $\D$ if ${({\A} \perp_d {\B}
  \mid {\C})}$ after deletion of every variable in $\D$. This is denoted \esep.
\end{definition}

Conditioning on $\C$ may close some paths between $\A$ and $\B$, and open
others. In the context of \(e\)-separation, the set $\D$, which we refer to as a
\textit{bottleneck} for $\A$ and $\B$ conditional on $\C$, is any set that
includes at least one variable from each path between $\A$ and $\B$ that is open
after conditioning on $\C$. If no subset of $\D$ is a bottleneck, then $\D$ is
called a \textit{minimal} bottleneck. This terminology reflects the fact that,
conditional on $\C$, all information shared between $\A$ and $\B$ -- that is,
transferred from one to the other or transferred to each from a common source --
must flow through $\D$.

It has been shown that every \(e\)-separation relationship among observed
variables in a graph $\mathcal G$ corresponds to a constraint on the observed
data distribution ${P(\Y)}$ \citep{Evans2012instrumental}. However, this result
is not constructive, in the sense that it does not provide a strategy for
deriving such constraints for a given \(e\)-separation relationship. The
inequality constraints we provide in Section~\ref{sec:constraints} partially
fulfill this role; they provide explicit constraints that hold everywhere in the
model whenever an \(e\)-separation relationship obtains in a graph.

\subsubsection{Node Splitting}

\begin{figure}
  \begin{center}
  \includegraphics{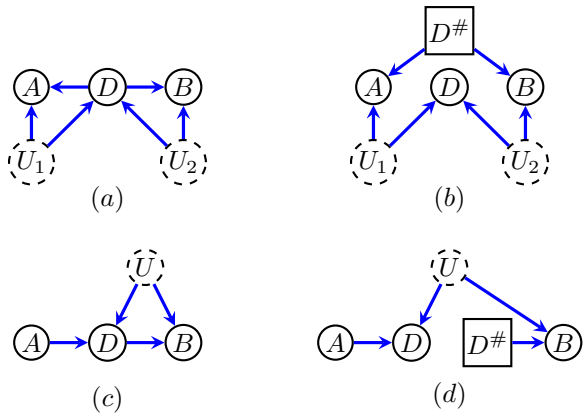}
  \end{center}
  \caption{
    The Unrelated Confounders graph (a), and a split node
    model for it (b), as well as the Instrumental graph (c), and its split node
    model (d).
  }
 \label{fig:node-splitting}
\end{figure}

We will see that \(e\)-separation is related to the idea of splitting
nodes in a graph. We define a node-splitting operation as follows. Given a graph
$\mathcal G$ and a vertex $D$ in the graph, the node splitting operation returns
a new graph $\GS$ in which $D$ is split into two vertices. One of the vertices
is still called $D$, and it maintains all edges directed into $D$ in the
original graph $\mathcal G$, but none of its outgoing edges. This vertex keeps
the name $D$ because it will have the same distribution as $D$ in the original
graph, as all of its causal parents remain the same. The second random variable
is labeled $D^\#$, and it inherits all of the edges outgoing from $D$ in the
original graph, but none of its incoming edges. Examples of the node
splitting operation are illustrated in Fig.~\ref{fig:node-splitting}.

By a result of~\citet{Evans2012instrumental}, ${(\A\perp_e \B\mid\C\; \text{~upon~}
\lnot \D})$ in $\G$ if and only if ${(\A\perp_d \B\mid\C,\DS)}$ in $\GS$. Note that
the node splitting operation described here is closely related to the operation of node splitting in Single World
Intervention Graphs in causal inference~\citep{Richardson2013SingleWI}.

\subsection{Entropies}

In this section, we review standard concepts in information theory, which we
will use to express our inequality constraints. We begin with the definitions of
entropy and mutual information.

\begin{definition}
  \label{def:entropy}
  The \textbf{entropy} of a random variable $X$ is defined as ${H(X) \equiv -
  \sum_{x \in \mathcal X} P(x) \log_2 P(x)}$, with the joint entropy of $X$ and
  $Y$ defined analogously. The \textbf{mutual information} between $X$ and $Y$
  is defined as ${I(X:Y) \equiv H(X) + H(Y) - H(X, Y)}$.
\end{definition}

The entropy of a random variable can be thought of as the level of uncertainty
one has about its value. Entropy is maximized by a uniform distribution over
the domain of a random variable, as there is no reason to think any one value is
more probable than another, and minimized by a point distribution, in which
there is no uncertainty.

The mutual information between $X$ and $Y$ can be thought of as the amount of
certainty we gain about the value of one, on average, if we learn the value of
the other. It is maximized when one of $X$ and $Y$ is a deterministic function
of the other, and is minimized when they are independent.

The entropy ${H(X \mid Y {=} y)}$ of $X$ conditional on a specific value of $Y {=} y$
is obtained by replacing the distribution ${P(X)}$ in Definition~\ref{def:entropy}
with ${P(X \mid Y {=} y)}$. The \textbf{conditional entropy} of $X$ given $Y$,
denoted ${H(X \mid Y)}$, is defined as the expected value of ${H(X \mid Y {=} y)}$.
Conditional mutual information is analogously defined.

\section{\(e\)-separation Constraints}
\label{sec:constraints}

We have already described the intuition behind our constraints, which can be
roughly summarized by the observation that the statistical dependence between
random variables must be limited by the total amount of information that can
flow through any bottleneck between them. We now describe how the tools
introduced in Section~\ref{sec:prelim} help us formalize this intuition.

First, we describe why \(e\)-separation helps formalize the idea of blocking
``all paths'' between two sets of variables. Consider the instrumental variable
graph, depicted in Fig.~\ref{fig:node-splitting}(c). $A$ and $B$ are only
\(d\)-separated by the set $\{D, U\}$, where $U$ is unobserved. Consequently,
they are not \(d\)-separated by any set consisting entirely of observed
variables. They are, however, \(e\)-separated after deletion of the observed
variable $D$. This tells us that all paths between $A$ and $B$ are through $D$,
and we can take advantage of observed properties of $D$ to bound the dependence
between them even when nothing is known about the unobserved variable $U$. A
similar story can be told about the Unrelated Confounders scenario depicted in
Fig.~\ref{fig:node-splitting}(a).

When all variables are observed, \(e\)-separation does not imply any constraints
that are not implied by \(d\)-separation, which follows from the fact that
\(d\)-separation implies all constraints in such cases~\citep{pearl1988reasoning}.
However, as illustrated by the examples in Figs.~\ref{fig:node-splitting}(a) and \ref{fig:node-splitting}(c), \(e\)-separation allows us to identify bottlenecks 
consisting entirely of observed variables between $A$ and $B$ even when paths between $A$ and $B$ cannot be closed by \emph{any} manner of conditioning on observed variables.
To show how \(e\)-separation lead to entropic constraints, we will make use of
Theorem 4.2 in \citep{Evans2012instrumental}, reframed as follows.

\begin{thm}\label{thm:evans}(\citet[Theorem 4.2]{Evans2012instrumental})\newline
  \noindent Suppose {\esep} in $\mathcal G$, and that no variable in $\C$ is a descendant of
  any in $\D$. Then there exists a distribution $P^*$ over $\A$, $\B$, $\C$, $\D$, $\DS$
  such that 
  \begin{align}\begin{split}\label{eq:counterfactual}
  &{P({\A{=}\a, \B{=}\b, \D{=}\d \mid \C{=}\c})} 
  \\&= {P^*({\A{=}\a, \B{=}\b,
    \D{=}\d \mid \C{=}\c, \DS {=}\d})}
\end{split}\end{align}
with ${\A \perp \B \mid \C, \DS}$ in $P^*$.
  If furthermore no variable in $\A$ is a descendant of any in $\D$, then there
  exists a distribution $P^*$ such that ${P({\B{=}\b, \D{=}\d \mid \A{=}\a,
    \C{=}\c})} ={ P^*({\B{=}\b, \D{=}\d \mid \A{=}\a, \C{=}\c, \DS {=}\d})}$ with
  ${\A \perp \B \mid \C, \DS}$ in $P^*$.
  \footnote{In causal inference problems, a distribution $P^*$ that satisfies the relevant conditions for this result may be constructed from counterfactual random variables ${\A}({\d}), {\B}({\d}), \D({\d})$ and ${\C}({\d})$.}
\end{thm}

We provide the following intuition for this theorem. Our graph $\G$ represents
the causal relationships within a system of random variables in the real world.
The graph $\GS$ represents an alternative world in which the causal effects of
$\D$ are ``spoofed'' by some random variable $\DS$. That is, children of $\D$ in
$\mathcal G$, which should be functions of $\D$, are instead fooled into being
functions of $\DS$.

In the alternative world represented by $\GS$, we suppose that the functional
form $f_V$ of a variable $V$ in terms of its parents stays the same for all
variables that are shared between graphs. This means that all non-descendants of
$\D$ have the same joint distribution in our world and in the alternative world,
as neither their parents nor the functions defining them in terms of their
parents have changed. By contrast, descendants of $\D$ in $\G$ will have a
different distribution in the alternative world, as their distributions are now
functions of the distribution of $\DS$, which may be different from that of
$\D$, and is unknown.

Now, suppose we condition on a particular value of ${\DS {=}\d}$ in $\GS$. Then,
because the functional form of the causal mechanisms is shared across worlds,
the descendants of $\D$ in $\mathcal G$ have the same distribution as they have
when  $\D {=}\d$ in the observed world. In addition, all of the non-descendants
of $\DS$ are marginally independent from $\DS$, because it has no ancestors so
all connecting paths must be collider paths. Therefore, both its non-descendants
and its descendants have the same joint distribution they would have had when
${\D{=}\d}$ in the original graph. The results in the theorem then follow when
we note that $\C$, and optionally $\A$, are non-descendants of $\D$, and that
the relevant independence properties hold in the world of $\GS$.

In general, we cannot know what this $P^*$ distribution is, because we never get
to observe this alternate world. But when we condition on $\DS$, we are
removing precisely the randomness we do not know about, yielding a distribution
that we do know about. The fact that $P^*$ agrees with $P$ on a subset of their
domains, and that it contains known independences, is sufficient to derive
informative constraints, as seen below.

\subsection{Entropic Constraints from \(e\)-separation}

We now show how the notion of \(e\)-separation permits the formulation of entropic
inequality constraints. In these constraints, we use mutual information to
represent dependence between sets of variables, and entropy to measure the
information-carrying capacity of paths connecting them.

\begin{thm}\label{thm:entropic}(Proof in Appendix~\ref{app:proofs}.)\newline
\noindent Suppose the variables in $\D$ are discrete. Whenever {\esep}, then ${I(\A:\B\mid\C,\D) \leq H(\D)}$. If in addition no element of $\C$ is a descendant of any in $\D$, then for any
value $\c$ in the domain of $\C$, the following stronger constraints hold:
\begin{subequations}\begin{align}
I(\A:\B\mid\C{=}\c,\D) &\leq H(\D\mid\C{=}\c)\label{eq:firststrong}\\
I(\A:\B\mid\C,\D) &\leq H(\D\mid \C).\label{eq:firstweak}
\end{align}\end{subequations}

If in addition, no element of $\A$ is a descendant of any in $\D$, then for any
value $\c$ in the domain of $\C$, the following even stronger constraints hold:
\begin{subequations}\begin{align}\label{eq:boundforMME}
I(\A:\B,\D\mid\C{=}\c) &\leq H(\D\mid\C{=}\c),\\
I(\A:\B,\D\mid\C) &\leq H(\D\mid \C).\label{eq:secondweak}
\end{align}\end{subequations}
\end{thm}

\begin{figure}[h!]
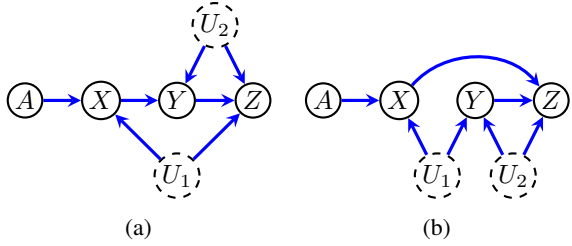

  \begin{center}
  \subcaptionbox{\label{fig:strongerconstrA}}
{\includegraphics{uai-figs/strongerconstrA}}
\hspace{3mm} 
\subcaptionbox{\label{fig:strongerconstrB}}
{\includegraphics{uai-figs/strongerconstrB}}
  \end{center}
  \caption{\label{fig:examplesoftheorem} For graph (a), Theorem~\ref{thm:entropic} implies the entropic inequality constrains ${I(A:XYZ)\leq H(X)}$ and ${I(A:YZ)\leq H(Y)}$. For graph (b), Theorem~\ref{thm:entropic} implies ${I(A:XYZ)\leq H(X)}$ and ${I(A:YZ\vert X)\leq H(Y\vert X)}$. Note, however, that the entropic quantifier elimination method of~\citet{Chaves2013Entropies} as applied by~\citet{Weilenmann2020}, finds that the former inequality for graph (b) can be strengthened into ${I(A:XYZ)\leq H(X\vert Y)}$.}
\end{figure}

This theorem potentially allows us to efficiently discover  many entropic inequalities implied by any given graph, such as those implied by Fig.~\ref{fig:examplesoftheorem}. In some cases, as in Fig.~\ref{fig:strongerconstrA}, the theorem recovers \emph{all} Shannon-type entropic inequality constraints implied by the graph  \citep{Chaves2013Entropies,Chaves2014Entropies,Weilenmann2017}. In other cases, as in Fig.~\ref{fig:strongerconstrB}, the graph implies a Shannon-type entropic inequality constraint beyond what Theorem~\ref{thm:entropic} can recover, per a result in \citep{Weilenmann2020}. Indeed, entropic inequality constraints can be implied by graphs not exhibiting \(e\)-separation relations whatsoever, such as the triangle scenario \citep{SteudelAy,Chaves2014Entropies}.

The linear quantifier elimination of \citep{Chaves2013Entropies,Chaves2014Entropies,Weilenmann2017} will always discover all the entropic inequalities which can be inferred from Theorem~\ref{thm:entropic}. However, the quantifier elimination method is computationally expensive, and is essentially intractable for graphs involving more than six or seven variables (observed and latent combined). Theorem~\ref{thm:entropic}, by contrast, provides an approach that is computationally tractable, but is capable of discovering fewer entropic constraints.

Finally, we note that an inverse of Theorem~\ref{thm:entropic} also holds. In particular, if ${(\A\not\perp_e \B\mid\C \text{ upon } \lnot \D)}$ in $\G$ and the variables in $\D$ are discrete, then there necessarily exists some distribution $P$ over $\A$, $\B$, $\C$, $\D$, in the marginal model of $\G$
such that ${I(\A:\B\mid\C,\D) \gneq H(\D)}$ when evaluated on $P$.

\subsection{Relating \(e\)-separation to Equality Constraints}
\label{sec:verma}

  We have seen that \(d\)-separation and \(e\)-separation relations imply
  constraints on the observed data distribution. \citet{verma1990verma} discuss
  equality constraints for latent variable models that apply in identified post-intervention distributions.  
  Such equality constraints are
  sometimes called Verma constraints.
  A general description of the class of these constraints implied by a hidden variable DAG model, as well as discussion of properties of these constraints is given in Ref.~\citep{richardson2017nested}.
  In this section, we examine the relationship between the \(e\)-separation-based constraints of Theorem \ref{thm:entropic} and the \(d\)-separation-based conditional independence and Verma constraints.

  First, we observe that the presence of \(d\)-separation relations and Verma
  constraints in a graphical model imply the presence of an \(e\)-separation
  relation.

\begin{prop}\label{prop:d-entails-e}(Proof in Appendix~\ref{app:proofs}.)\newline
  \noindent If $\A$ is \(d\)-separated from $\B$ by $\{\C,\D\}$, then $\A$ is also $e$-separated
  from $\B$ by $\C$ upon deleting $\D$.
\end{prop}

This demonstrates that for any \(d\)-separation relation in the graph, it is
possible to obtain an entropic constraint corresponding to any minimal
bottleneck $\D$ through an \(e\)-separation relation. More precisely, when $\A$
is $d$-separated from $\B$ by $\{\C, \D\}$, then by
Proposition~\ref{prop:d-entails-e}, it is also the case that $\A$ is
$e$-separated from $\B$ given $\C$ upon deleting $\D$, and therefore
Theorem~\ref{thm:entropic} can be applied to obtain entropic constraints. Note,
however, that these are necessarily weaker than the entropic constraint
${I(\A:\B\mid\C,\D)=0}$, which follows from the $d$-separation relation itself.

In summary, every \(d\)-separation relation in the graph is an instance of \(e\)-separation,
but not vice-versa. When an instance of \(e\)-separation is also an instance of \(d\)-separation,
then all the inequality constraints implied by \(e\)-separation are rendered defunct by the stronger equality constraints
implied by \(d\)-separation.

We now show that a similar pattern of deprecating inequalities by equalities occurs in the presence of Verma constraints when certain counterfactual interventions are identifiable.

\begin{prop} \label{prop:stronger_entropic}
  Consider a graph  $\mathcal G$ which exhibits the \(e\)-separation relation
  {\esep} and where no element of $\C$ is a descendant of any in $\D$. If the counterfactual distribution ${P(\A{(\D{=}\d)},\B{(\D{=}\d)},\D{(\D{=}\d)} \mid \C)}$ is identifiable\footnote{The counterfactual distribution in this theorem allows intervened-on variables and outcomes to intersect.  See \citep{shpitser21multivariate} for a complete identification algorithm for counterfactual distributions of this type.} then the inequalities of Theorem~\ref{thm:entropic} are logically implied whenever the stronger equality constraints
  \begin{align}\label{eq:entropicequality}
  I\left(\A(\D{=}\d):\B(\D{=}\d)\mid \C\right)=0
  \end{align}
  are satisfied for all values of $\d$. Note that Equation~\eqref{eq:entropicequality} is satisfied if and only if the \emph{margin} of the identified counterfactual distribution factorizes, i.e., when 
  \begin{align}
  &{P(\A{(\D{=}\d)},\B{(\D{=}\d)} \mid \C)}\nonumber
  \\&\equiv {\textstyle\sum\nolimits_{\d'} P(\A{(\D{=}\d)},\B{(\D{=}\d)},\D(\D{=}\d){=}\d' \mid \C )}\nonumber
   \\&\qquad\text{exhibits}\quad \A{(\D{=}\d)}\perp \B{(\D{=}\d)}\mid \C.\label{eq:kernel_equality} 
  \end{align}
  \end{prop}
  
The proof directly follows from that of Theorem~\ref{thm:entropic}. In proving
Theorem~\ref{thm:entropic}, we derive entropic inequalities by relating the entropies pertaining to $P(\A,\B,\D\mid\C)$ to entropies pertaining to the $P^*$ distribution posited by Theorem~\ref{thm:evans}. That is, Theorem~\ref{thm:entropic} is an entropic consequence of Theorem~\ref{thm:evans}. If the conditions of Proposition~\ref{prop:stronger_entropic} are satisfied, then the conditions of Theorem~\ref{thm:evans} are also automatically satisfied since one can then \emph{explicitly} reconstruct 
\begin{align}\begin{split}\label{eq:Pstarmap}
&P^*({\A, \B,\D{=}\d \mid \C, \DS {=}\ds})
    \\&= P(\A{(\D{=}\ds)},\B{(\D{=}\ds)},\D(\D{=}\ds){=}\d \mid \C ).
\end{split}\end{align}
There is no opportunity to violate the entropic inequalities of Theorem~\ref{thm:entropic} once the observational data has been confirmed as consistent with Theorem~\ref{thm:evans}. In other words, in order to violate the inequalities of Theorem~\ref{thm:entropic} it must be the case that no $P^*$ consistent with Theorem~\ref{thm:evans} can be constructed, but this contradicts the explicit recipe of Equation~\eqref{eq:Pstarmap}.

See Refs.~\citep{verma1990verma, tian2002testable, richardson2017nested} for details
on how to derive the form of the equality constraints summarized by
Equation~\eqref{eq:kernel_equality}. We note here that
${P(\A{(\D{=}\d)},\B{(\D{=}\d)},\D(\D{=}\d){=}\d \mid \C )}$ is certainly
identifiable if $\D$ is not a member of the same \emph{district}~(\citep{richardson2017nested}) as any element
in $\{\A,\B\}$ within the subgraph of $\mathcal{G}$ over $\{\A,\B,\C,\D\}$ and
their ancestors. We also note that the identifiability of \emph{merely}
${P(\A{(\D{=}\d)},\B{(\D{=}\d)}\mid \C )}$ but not of
${P(\A{(\D{=}\d)},\B{(\D{=}\d)},\D(\D = \d){=}\d\mid \C )}$ negates the
implication from Equation~\eqref{eq:kernel_equality} to
Theorem~\ref{thm:entropic}. In Appendix~\ref{app:verma}, we provide an example
of a graph in which ${P(\A{(\D{=}\d)},\B{(\D{=}\d)}\mid \C )}$ is identified, but
the entropic constraints of Theorem~\ref{thm:entropic} remain relevant. In
addition, we demonstrate that the application of the entropic constraints to
identified counterfactual distributions can also result in inequality
constraints on the observed data distribution.

\subsection{Constraints and Bounds Involving Latent Variables}

In this section, we consider \(d\)-separation relations with hidden variables in
the conditioning set. Because we cannot condition on hidden variables, there is
no way to check whether the corresponding independence constraints hold in the
full data distribution. However, if we have access to auxiliary information
about these hidden variables -- such as information about their entropy or their
cardinality -- it is possible to obtain inequality constraints on the observed
data distribution.

\begin{prop}\label{prop:entropic-dsep}(Proof in Appendix~\ref{app:proofs}.)\newline
\noindent If  ${(\A \perp_d \B \mid \C, \Ub)}$ and ${{H(\Ub \mid \A,\B,\C{=}\c)}\geq 0}$, for any
value $\c$ in the domain of $\C$:
\begin{align}
{H({\Ub \mid \C{=}\c})} &\ge {I({\A : \B \mid \C{=}\c})}
\end{align}

\end{prop}
Note that Proposition~\ref{prop:entropic-dsep} holds even if $\A$, $\B$, and $\C$ are continuously valued variables.

In many scenarios, we may have more (or be more interested in) information pertaining to the \textit{cardinality}
of a hidden variable than its entropy. We take the cardinality of a set of
variables to be the product of the cardinalities of the variables in the set. An
upper bound on the cardinality of $\Ub$ entails an upper bound on its entropy.
As observed above, the entropy of a random variable is maximized when it takes a
uniform distribution. If we let $|\Ub|$ denote the cardinality of $\Ub$, and
recall that the entropy of a uniformly distributed variable with cardinality $m$
is simply $\log_2 (m)$, then $\log_2 |\Ub| \ge H(\Ub)$. The next corollary then
follows immediately from Proposition~\ref{prop:entropic-dsep}, since $H(\Ub\mid\cdot)\geq 0$ whenever $\Ub$ has finite cardinality:

\begin{samepage}\begin{cor}\label{cor:latent-cardinality}
  If ${(\A \perp_d \B \mid \C, \Ub)}$, then for any
value $\c$ in the domain of $\C$, the cardinality of $\Ub$ may be lower-bounded:
\begin{align}
|\Ub| \geq \max\nolimits_\c \; 2^{I({\A : \B \mid \C{=}\c})} \geq 2^{I({\A : \B \mid \C})}.
\end{align}
\end{cor}
\end{samepage}

Finally, we note that both of these inequalities can also be used if we
\textit{do not} know anything about the properties of $\Ub$, but would like to
infer lower bounds for its entropy and cardinality from the observed data. In Section
\ref{sec:genetics}, we will explore a scenario in genetics in which these bounds
and constraints may be of use.

\begin{remark}\label{remark:d-sep}
Constraints given in Proposition~\ref{prop:entropic-dsep} and Corollary~\ref{cor:latent-cardinality} are stronger than can be obtained from the \mbox{\(e\)-separation}
relation ${(\A \perp_e \B \mid \C \text{~upon~} \lnot \Ub)}$ on its own. 
\end{remark}
To demonstrate Remark~\ref{remark:d-sep}, we consider a set of structural equations consistent
with Fig.~\mbox{\ref{fig:node-splitting}(a)}. Suppose that $D$ takes the value $0$
when $U_1 \neq U_2$, and the value $1$ otherwise, and that $A$ and $B$ take the
value $0$ if $D$ is $0$, and values equal to $U_1$ and $U_2$ respectively if $D$
is $1$. It follows that $A$ and $B$ are always equal, and therefore ${I(A:B)=H(A)}$. Now, suppose that $U_1$
and $U_2$ only take values not equal to $0$, and that there are at least two
values that each takes with nonzero probability. It immediately follows that
${H(D) < H(A)}$, and therefore that ${H(D) < I(A:B)}$, as $D$ and $A$ by
construction take the value $0$ with the same probability, but there is strictly
more entropy in the remainder of $A$'s distribution because $D$ is binary and
$A$ takes at least two other values with nonzero probability.

\section{Applications}
\label{sec:applications}

In this section, we explore several applications of the constraints developed above. In Sections~\ref{sec:discovery} and~\ref{sec:genetics}, we show how our results can be used to learn about causal models from observational data. In Section~\ref{sec:mme}, we further leverage the importance of the entropy of variables along a causal pathway to posit a new measure of causal strength, and observe that this measure can be bounded by an application of Theorem~\ref{thm:entropic}.

\subsection{Causal Discovery}
\label{sec:discovery}

\begin{figure}[tb!]
  \begin{center}
  \includegraphics{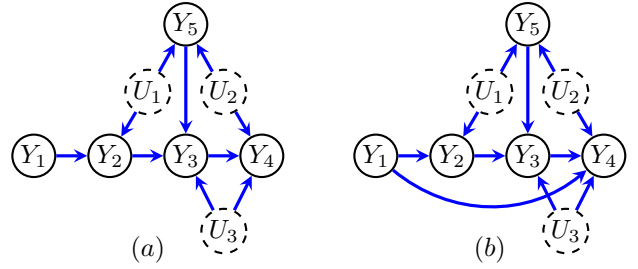}
  \end{center}
  \caption{
    Two hidden variable DAGs that share equality constraints over
    observed variables, but (a) contains \(e\)-separation relations that are not in (b).
  }
 \label{fig:discovery}
\end{figure}

In this section, we present an example in which two hidden variable DAGs with
the same equality constraints present different entropic inequality constraints.
The ability to distinguish between models that share equality constraints has
the potential to advance the field of causal discovery, in which causal DAGs are
learned directly from the observed data. Causal discovery algorithms for
learning hidden variable DAGs currently do so using only equality constraints.
Our approach may be useful as a post-processing addition to such methods,
whereby any graph found to satisfy the equality constraints in the observed data
is tested against the entropic inequality constraints implied by
\(e\)-separation relations in the model.

The hidden variable DAGs in Fig.~\ref{fig:discovery}, adapted from
Appendix~B in Ref.~\citep{bhattacharya2020differentiable}, share the same conditional independence
constraints: $Y_1 \perp Y_3 \mid Y_2 Y_5$ and $Y_1 \perp Y_5$, but
exhibit different \(e\)-separation relations. 

In Fig.~\mbox{\ref{fig:discovery}(a)}, ${(Y_1
\perp_e Y_3Y_4 \mid Y_2 \text{~upon~} \lnot Y_5)}$, ${(Y_1Y_2 \perp_e Y_4 \mid
\text{~upon~} \lnot Y_3)}$, and ${(Y_2 \perp_e Y_4 \mid Y_1 \text{~upon~} \lnot
Y_3)}$. Applying Theorem~\ref{thm:entropic} in each case, we obtain the three inequality
constraints ${I(Y_1 \!:\! Y_3Y_4Y_5 \mid Y_2) \le H(Y_5 \mid Y_2)}$, ${I(Y_2 \!:\! Y_3 Y_4 \mid Y_1) \le H(Y_3 \mid Y_1)}$, ${I(Y_1Y_2 \!:\! Y_3Y_4) \le H(Y_3)}$.

In Fig.~\mbox{\ref{fig:discovery}(b)}, we have added an edge, which removes some
\(e\)-separation relations. We are left with ${(Y_1 \perp_e Y_3 \mid Y_2
\text{~upon~}\lnot Y_5)}$, and ${(Y_2 \perp_e Y_4 \mid Y_1 \text{~upon~} \lnot
Y_3)}$. We can again apply Theorem~\ref{thm:entropic} in each case, yielding the inequality constraints ${I(Y_1 : Y_3 Y_5 \mid Y_2)
\le H(Y_5 \mid Y_2)}$ and ${I(Y_2 : Y_3 Y_4 \mid Y_1) \le H(Y_3 \mid Y_1)}$. The
second of these constraints is shared by the graph in
Fig.~\mbox{\ref{fig:discovery}(a)}, and the first is strictly weaker than a constraint
in Fig.~\mbox{\ref{fig:discovery}(a)}.

Models similar to those shown in Fig.~\ref{fig:discovery} sometimes arise in
time-series data, where the variables in the chain represent observations taken
at consecutive time steps. In such models, it is often assumed that treatments
no longer have a direct effect on outcomes after a certain number of time steps.
Here, that assumption is encoded in the lack of a direct edge from $Y_1$ to
$Y_4$ in Fig.~\mbox{\ref{fig:discovery}(a)}. We have shown above that this kind of
assumption can be falsified even when it does not imply any additional equality
constraints, as is often the case. In particular, if the stronger constraints
implied by Fig.~\mbox{\ref{fig:discovery}(a)} are violated, but the weaker constraints
of Fig.~\mbox{\ref{fig:discovery}(b)} are not, then the assumption is falsified.

\subsection{Causal Discovery in the Presence of Latent Variables}
\label{sec:genetics}

\begin{figure}[h]
  \begin{center}
  \includegraphics{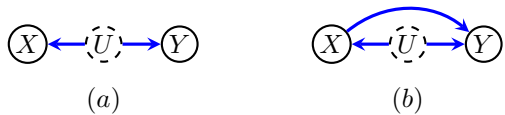}
  \end{center}
  \caption{
    Identifying direct causal influence in the presence of a confounder with limited cardinality.
  }
  \label{fig:identifying}
\end{figure}

In this section, we consider a very simple possible application of the
constraints and bounds relating to entropies of unobserved variables in
genetics. Consider a causal hypothesis wherein the presence or absence of an
unobserved gene influences two aspects of an organism’s phenotype. Suppose
that due to genetic sequencing studies, the number of variants of the gene in
the population -- i.e. the cardinality of the corresponding random variable --
is known. Two possible hypotheses regarding the causal structure
are depicted in Fig.~\ref{fig:identifying}, where $U$ represents the gene and
$X$ and $Y$ are the phenotype aspects.  In Fig.~\mbox{\ref{fig:identifying}(a)}, one
presumes no causal influence of $X$ on $Y$, whereas in
Fig.~\mbox{\ref{fig:identifying}(b)}, direct causal influence is allowed. In the former case, knowledge of the number of variants of the gene constrains the mutual information
between the phenotypes, while in the latter case it is not constrained. 

Thus, for certain types of statistical dependencies between $X$ and $Y$, one can rule
out the hypothesis of Fig.~\mbox{\ref{fig:identifying}(a)}. For example, suppose we
know the cardinality of $U$ to be $3$. Corollary~\ref{cor:latent-cardinality}
then implies the constraint that the mutual information between $X$ and $Y$
cannot exceed $\log_2(3) \approx 1.584$. Suppose further that we observe the
distribution depicted in Table~\ref{tab:dist}. The mutual information between
$X$ and $Y$ in this distribution is $\approx 1.594$. Because this mutual
information violates the constraint implied by the model in
Fig.~\mbox{\ref{fig:identifying}(a)}, we know this model cannot be correct, and
conclude that Fig.~\mbox{\ref{fig:identifying}(b)} is correct. More generally, strong
statistical dependence between high cardinality variables cannot be explained by
a low cardinality common cause and requires a direct influence between them.

\begin{table}[h]
\begin{center}
\begin{tabular}{ c c | c c c c }
 & & & \quad\quad\quad Y & & \\
 & & 0 & 1 & 2 & 3 \\\hline
 & 0 & 0.002 & 0.001 & 0.400  & 0.001\\
 X & 1 & 0.003 & 0.005 & 0.005 & 0.066\\
 & 2 & 0.224 & 0.003 & 0.003 & 0.001\\
 & 3 & 0.002 & 0.281 & 0.001 & 0.002\\
\end{tabular}
\end{center}
\caption{An example joint distribution over two variables $X$ and $Y$, each with
  cardinality $4$.
}
\label{tab:dist}
\end{table}

Conversely, suppose Fig.~\mbox{\ref{fig:identifying}(a)} is known to be correct, and
that there is no direct causal influence between the two aspects of phenotype.
If the cardinality of $U$ is not known, it can be bounded from below directly
from observed data, according to Corollary~\ref{cor:latent-cardinality}. In this
case, the lower bound would be $2^{I(X:Y)} \approx 2^{1.594} \approx 3.018$. It
follows that $U$ must have a cardinality of $4$ or above in this setting. The
ability to extract such information from observational data may be useful in
making substantive scientific decisions, or in guiding future sequencing
studies.

In many applied data analyses, different variables may be observed for different
subjects, i.e., data on some variables is ``missing'' for some subjects. A recent
line of work has focused on properties of missing data models that can be
represented as DAGs \citep{mohan2013missing-dag}. Although the bounds and constraints
above have been developed in the context of fully unobserved variables, they can
also be used in missing data DAG models, for variables that are not observed for
all subjects.

\subsection{Quantifying Causal Influence}
\label{sec:mme}

The traditional approach to measuring the strength of a causal relationship is
by contrasting how different an outcome would be, on average, under two
different treatments. Formally, if $X$ is a cause of $Y$, the \ACE is defined as
$E[Y(X=x) - Y(X=x')]$. While the \ACE is a very useful construct, we suggest
that it has two important shortcomings, and present an alternative measure of
causal strength called the \emph{Minimal Mediary Entropy} or \emph{\MME}. The \MME is based on the idea -- explored throughout this work -- that the entropy of variables along a causal pathway provide insight into the amount of information that can travel along that pathway.

In a scenario where treatment can be discerned to always cause outcome, we might expect the \ACE, as a measure of causal influence, to be large. The example below shows that this is not necessarily the case.

\begin{example}
  \label{ex:ace}
  Consider a randomized binary treatment $X$ and a ternary outcome $Y$, with
  ${P(Y {=} 0 \mid X {=} 0)} = {P(Y {=} 2 \mid X {=} 0)} = 0.5$, and ${P(Y {=} 1 \mid X {=} 1)} =
  1$. In this setting, $\ACE = 0$, even though treatment affects outcome
  for every subject in the population.
\end{example}

In less extreme settings, the \ACE may be very small even when
treatment affects outcome for almost every subject in the population, or very large, even when very few subjects have an outcome that is affected
by treatment.

The \ACE is likewise not always well suited to measuring the
strength of a causal relationship when the treatment variable is non-binary. In such
situations, no one causal contrast represents the causal influence, and
the number of possible contrasts grows combinatorially in the cardinality of
treatment. We now define the \MME and discuss how it can overcome these issues.

\begin{figure}[htb!]
  \begin{center}
  \includegraphics{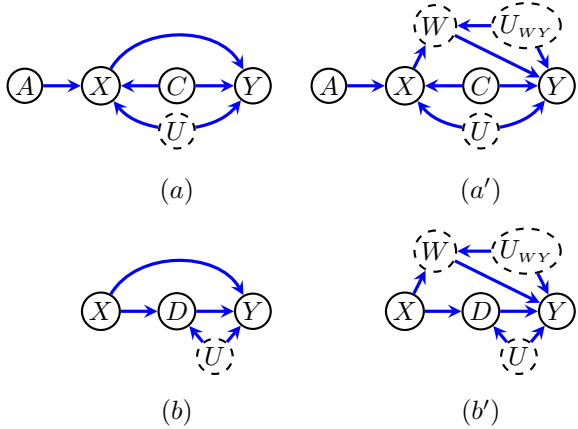}
  \end{center}
  \caption{
    Modifying DAGs (a) and (b) by inserting a latent mediary $W$ between $X$ and $Y$ yields DAGs (a') and (b') respectively. Note that in (a), even though $X$ and $Y$ are latent confounded, corollary~\ref{cor:mme-bound} gives ${\MME_{X \to Y}\geq I(A:Y\mid C)}$ by exploiting the fact that $A\in an(X)$. Also note that in (b), even though $X$ affects $Y$ both directly and indirectly though $D$, corollary~\ref{cor:mme-bound} gives ${\MME_{X \to Y}\geq I(X:Y\mid D)-H(D)}$ for the direct effect.
  }
 \label{fig:mmc}
\end{figure}

\begin{definition}[Minimal Mediary Entropy (\MME) for Direct Effect]\label{def:MME}
Given a DAG $\G$ containing a directed edge ${X\to Y}$, let $\G^\prime_{X\to W\to Y}$ denote the graph constructed by substituting the single edge ${X\to Y}$ in $\G$ with the set of four edges ${\lbrace {X\to W\to Y},\;\;\;\;{W\leftarrow U_{WY} \to  Y}\rbrace}$, introducing auxilliary latent variables  $W$ and $U_{WY}$.\footnote{
    If a latent confounder were added between $X$ and $W$, then although $W$ would still mediate ${X \to Y}$, $X$ and $Y$ would share a source of unobserved confounding, altering the causal model.
 }
 We then define $\MME_{X\to Y}$ as the
  smallest entropy $H(W)$ over all structural equations models reproducing the observed data
  distribution over $\G^\prime_{X\to W\to Y}$ in which $W$ has finite cardinality.
\end{definition}
Fig.~\ref{fig:mmc} illustrates the process of edge substitution. Essentially, the edge ${X\to Y}$ in $\G$ is interrupted to pass through $W$ in $\G^\prime_{X\to W\to Y}$, such that the auxiliary latent variable $W$ \emph{fully mediates} the direct effect of $X$ on $Y$.
Note that caveat that ${\MME}$ is defined in terms of minimizing the entropy of $W$ over \emph{finite cardinality} $W$ capable of reproducing the observed statistics. If $W$ were allowed to be a continuously valued variable, then the observed data distribution would always be reproducible with arbitrarily small $H(W)$, due to the total lack of restriction in the instrumental model with a continuous mediary \citep{gunsilius2020pathsampling}.

With the presumption of finite cardinality $W$ by fiat, however, we are are in a position to exploit Theorem~\ref{thm:entropic} in order to practically lower bound the ${\MME}$.

\begin{cor} \label{cor:mme-bound}(Proof in Appendix~\ref{app:proofs}.)\newline
  Suppose that graphical construction $\G^\prime_{X\to W\to Y}$ exhibits the \(e\)-separation relation ${(\A\perp_e \B\mid\C \text{ upon } \lnot \{\D,W\})}$ and furthermore no element of $\{\A,\C\}$ is a descendant of any in $\{\D,W\}$, %
  where $\A$, $\B$, $\C$, and $\D$ are nonoverlapping subsets ($\C$ and $\D$ possibly empty) of the observed variables in $\G$, and all the variables within $\D$ are discrete. Then 
 \begin{align}
  \hspace{-1em}\MME_{X \to Y}\geq I(\A : \B \mid \C,\D)-H(\D\mid \C)\,.\label{eq:MMEaverageCwithD}
  \end{align}
\end{cor}

\color{black}

Suppose that $P_0$ is a distribution in the model of the extended graph  $\G^\prime_{X\to W\to Y}$, such that $P_0$ marginalizes to the observed data distribution. Then the entropy $H(W)$ in $P_0$ is necessarily an upper bound on $\MME_{X \to Y}$, i.e. we have found a $W$ with entropy $H(W)$ that fully mediates the causal influence of $X$ on $Y$. Since $W$ could always reproduce the observed data by simply copying the values of $X$, we have a trivial upper bound of $\MME_{X \to Y}\leq H(X)$.\footnote{Consider example~\ref{ex:ace} which has ${\ACE_{X\to Y}=0}$. That example has the feature that $I(X:Y)=H(X)$. Accordingly the lower bound of  ${\MME_{X \to Y}\geq I(X:Y)}$ is evidently tight, given the trivial upper bounder ${\MME_{X \to Y}\leq H(X)}$.} This upper bound can typically be improved by even partially exploring the space of the distributions in $\G^\prime_{X\to W\to Y}$.

Consider the simple model ${X \to Y}$ with ${|X| = 3}$ and ${|Y| = 3}$, and the observed data distribution %
$P(X{=}x, Y{=}y)=\begin{cases}\frac{5}{27}\text{ if }x=y\\\frac{2}{27}\text{ if }x\neq y \end{cases}$.
Our corollary gives us a lower bound on ${\MME_{X \to Y}\geq I(X:Y)\approx 0.150}$, contrasted with the trivial upper bound  ${\MME_{X \to Y}\leq H(X)\approx 1.585}$. We can improve the trivial upper bounding by noting that this distribution can be reproduced %
by the following functional relationships $\begin{cases} W{=}0\text{ and }Y{=}U_{WY}& \text{when }U_{WY}{=}X \\ W{=}1\text{ and }Y{=}\text{uniformly random}& \text{when }U_{WY}{\neq}X   \end{cases}$ and taking $U_{WY}$ to be a uniform random distribution with cardinality three.  %
In this model for  $\G^\prime_{X\to W\to Y}$ we obtain ${P(W{ = }0) = \frac{1}{3}}$, corresponding to ${\MME_{X \to Y}\leq H(W)\approx 0.918}$.

\section{Related Work}
\label{sec:related-work}
This work builds most directly on Ref.~\citep{Evans2012instrumental}, in which
\(e\)-separation was introduced and Theorem~\ref{thm:evans} was derived, both of
which are essential to our results. It follows in the tradition of a line
of literature that aims to derive symbolic expressions of restrictions on the
observed data distribution implied by a causal model with latent variables,
including Refs.~\citep{tian2002testable, balke1993ivbounds,KangTian2006}.
Entropic constraints were previously considered in Refs.~\citep{Chaves2013Entropies, Chaves2014Entropies,Weilenmann2017}. The
entropic constraint for the instrumental scenario appears as Equation~(5) in Ref.~\citep{Chaves2014Entropies}, see also Appendix~E of Ref.~\citep{Henson2014}.
Our work is also closely related to work in the literature on information theory
on how much information can pass through channels of varying
types~\citep{gamal2011communication}. Our proposed measure of causal strength,
the \MME, is motivated by weaknesses in standard causal strength measures (e.g.
ACE), which was previously discussed in Ref.~\citep{janzing2013quantifying}.

Our results are also related to the causal discovery literature, which seeks to
find the causal structures compatible with an observed data
distribution~\citep{spirtes2000causation}. The inequality constraints posed
above can be used to check the outputs of existing causal discovery
algorithms~\citep{strobl2018fast,
  spirtes2000causation, bernstein2019discovery}.
\section{Conclusion}
\label{sec:conclusion}
In this work, we present inequality constraints implied by \(e\)-separation
relations in hidden variable DAGs. We have shown that these constraints can be
used for a number of purposes, including adjudicating between causal models,
bounding the cardinalities of latent variables, and measuring the strength of a
causal relationship. \(e\)-separation relations can be read directly off a hidden
variable DAG, leading to constraints that can be easily obtained.

This work opens up two avenues for future work. The first is that our
constraints demonstrate a practical use of \(e\)-separation relations, and should
motivate the study of fast algorithms for enumerating all such relations in
hidden variable DAGs. The second is that the constraints
suggest that equality-constraint-based causal discovery algorithms can
be improved; understanding how the inequality constraints can best be used to
this end will take careful study.

\subsubsection*{Acknowledgments}

This research was supported by Perimeter Institute for Theoretical Physics.
Research at Perimeter Institute is supported in part by the Government of Canada
through the Department of Innovation, Science and Economic Development Canada
and by the Province of Ontario through the Ministry of Colleges and
Universities.
The first author was supported in part by the Mathematical Institute for Data Science (MINDS) research fellowship.
The third author was supported in part by grants ONR
N00014-18-1-2760, NSF CAREER 1942239, NSF 1939675, and R01 AI127271-01A1. We thank Murat Kocaoglu for helpful discussion about the ${\MME}$. We thank Robin Evans for motivating us to revise the proof of Theorem~\ref{thm:entropic} in Appendix~\ref{app:proofs}.

\onecolumn
\setlength{\bibsep}{3pt plus 3pt minus 2pt}
\nocite{apsrev42Control}
\bibliography{Finkelstein_392}

\appendix
\renewcommand{\theequation}{S\arabic{equation}}
\renewcommand\thefigure{S\arabic{figure}}

\section{Comparing Entropic Inequalities to Generalized Independence Relations}
\label{app:verma}

\begin{figure}[h!]
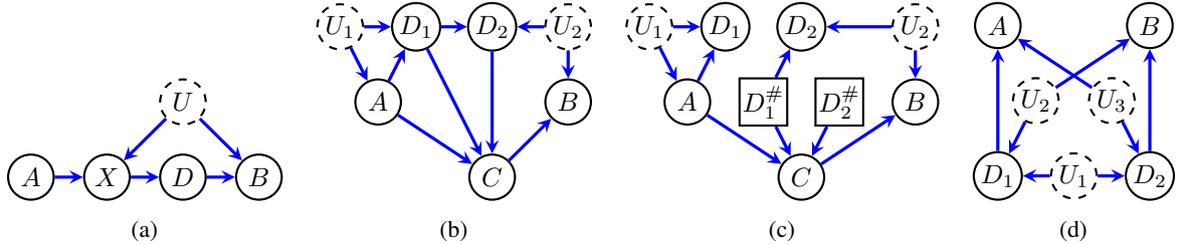

  \begin{center}
  \subcaptionbox{\label{fig:appendixA}}
{\includegraphics{uai-figs/sup1A}}
\hspace{3mm} 
\subcaptionbox{\label{fig:appendixB}}
{\includegraphics{uai-figs/sup1B}}
\hspace{3mm} 
\subcaptionbox{\label{fig:appendixC}}
{\includegraphics{uai-figs/sup1C}}
\hspace{3mm} 
\subcaptionbox{\label{fig:appendixD}}
{\includegraphics{uai-figs/sup1D}}

  \end{center}
  \caption{In graphs (a) and (b), the entropic  inequality constraints are logically implied by equality constraints. Graphs (b) and (c) demonstrate that for a set of variables $\D$, the counterfactual random variable $\D(\D=\d)$ is not necessarily equal to the factual $\D$. Graph (d) provides an example where the entropic inequality constraints remain relevant even though the counterfactual distribution after intervention on an \(e\)-separating set over the remaining variables is identified.}
\end{figure}

\sloppy In Proposition~\ref{prop:stronger_entropic}, we showed that for graphical models in which the counterfactual ${P(\A{(\D{=}\d)},\B{(\D{=}\d)},\D(\D{=}\d){=}\d \mid \C )}$ is identified, the entropic constraints of Theorem~\ref{thm:entropic} are weaker than the corresponding Verma constraints. We now illustrate this point with a few examples.
In Fig.~\ref{fig:appendixA}  the counterfactual $P(\A{(\D{=}\d)},\B{(\D{=}\d)},\D(\D{=}\d){=}\d)$ is identified, and in Fig.~\ref{fig:appendixB}  the counterfactual $P(\A{(\D{=}\d)},\B{(\D{=}\d)},\D(\D{=}\d){=}\d \mid \C)$ is identified. Accordingly, our entropic inequalities are implied by equality constraints, due to Proposition \ref{prop:stronger_entropic}. The resulting inequality constraints therefore cannot provide any additional information about whether these causal structures are compatible with observed distributions.

By contrast, in Fig.~\ref{fig:appendixD} the counterfactual $P(\A{(\D{=}\d)},\B{(\D{=}\d)},\D(\D{=}\d){=}\d \mid \C )$ is not identified, even though $P(\A{(\D{=}\d)},\B{(\D{=}\d)}\mid \C )$ is. Although Fig.~\ref{fig:appendixD}  implies no equality constraints \citep{NestedMarkovCounting}, we find that it \emph{does} entail the entropic inequality constraint following from the \(e\)-separation relation $(A \perp_e B \mid \text{~upon~} \lnot \{D_1, D_2\})$. It is therefore an example of a graph in which our inequality constraints are \emph{not} made redundant by known equality constraints, despite the fact that intervention on $\D$ is identified. This example is also an illustration of the fact that not every equality restriction featuring non-adjacent variables in an identifiable counterfactual distribution implies equality restrictions on the observed data distribution.  However, some such non-adjacent variables may be involved in inequality restrictions.

The critical idenfifiability question for determining whether the entropic constraints are made redundant by equality constraints is $P(\A{(\D{=}\d)},\B{(\D{=}\d)},\D(\D{=}\d){=}\d \mid \C )$. This distribution involves the counterfactual random variable $\D(\D{=}\d)$. Note that although any \emph{single} random variable under intervention on itself is equivalent to the random variable under no intervention, the same does not necessarily hold for \emph{sets} of random variables. Figs.~\ref{fig:appendixB}~and~\ref{fig:appendixC} demonstrate this point -- because $D_2$ is a descendant of $D_1$, after intervention on both, $D_2$ no longer takes its natural value.

\clearpage
\section{\(e\)-Separation in Identified Counterfactual Distributions}
\label{app:nesting}

\begin{figure}[h!]
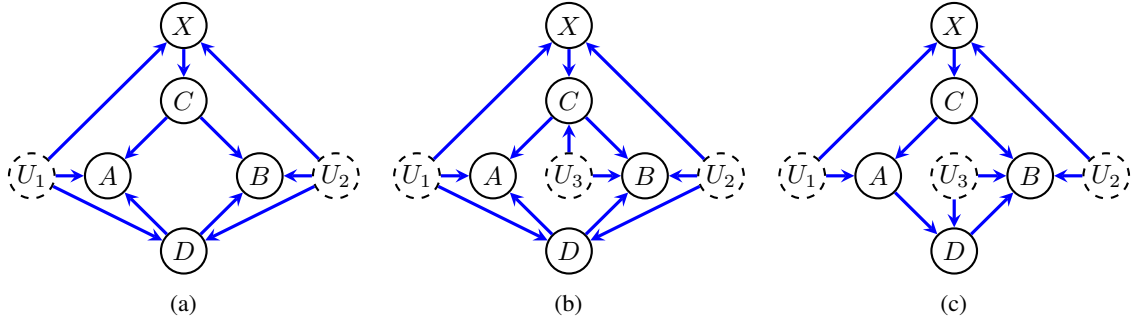

  \begin{center}
  \subcaptionbox{\label{fig:appendix2A}}
{\includegraphics{uai-figs/sup2A}}
\hspace{3mm} 
\subcaptionbox{\label{fig:appendix2B}}
{\includegraphics{uai-figs/sup2B}}
\hspace{3mm} 
\subcaptionbox{\label{fig:appendix2C}}
{\includegraphics{uai-figs/sup2C}}
  \end{center}
  \caption{In all three graphs, $A$ and $B$ are \(e\)-separated by $D$ after intervention on $C$. The counterfactural distribution over $\{A,B,D\}$ after intervention on $C$ is only identified in graphs (a) and (c), however.}
  \label{fig:nested-e-sep}
\end{figure}

    A Single World Intervention Graph (SWIG) \citep{Richardson2013SingleWI}, which represents the model after intervention on one or more random variables, can be obtained through a node-splitting operation as illustrated in Fig.~\ref{fig:appendixC}. As described in Section \ref{sec:verma}, \(d\)-separation relations that appear under interventions with identified distributions can be used to derive equality constraints on the observed data distribution. In this section, we explore the significance of \(e\)-separation relations in identified counterfactual distributions.
  
    We begin by noting that any \(e\)-separation relation that exists in a SWIG corresponds to an \(e\)-separation in the original DAG.
    
    \begin{prop}
      \esep~ after intervention on $\boldsymbol E$ only if $(\A \perp_e \B \mid \C \text{~upon~} \lnot \{\D, \boldsymbol E\})$.
    \end{prop}

    This proposition follows directly from the relationship between the fixing \citep{richardson2017nested} and deletion operations. In particular, fixing and deleting vertices induce the same graphical relationships among the remaining variables in the graph.
    
    It may at first seem that this result indicates that \(e\)-separation relations in SWIGs cannot be used to derive inequality constraints on the observed data distribution that are not already implied by \(e\)-separation relations in the original model. However, entropic inequality constraints on counterfactual distributions have a different form than such constraints on the factual distribution. This is because entropies of counterfactual variables do not in general correspond to entropies of factual variables, so there is no way to express inequality constraints that follow from \(e\)-separation relations in SWIGs as entropic inequalities on the original distribution.

    To illustrate this point, consider Fig.~\ref{fig:nested-e-sep}. In each graph, $(A \perp_e B \mid \text{~upon~} \lnot D)$ in the SWIG resulting from intervention on $C$. However, in Fig.~\ref{fig:appendix2B}, the distribution after intervention on $C$ is not identified, whereas in Figs.~\ref{fig:appendix2A} and \ref{fig:appendix2C} it is identified as $P(A(c), B(c), D(c)) = \sum_x \frac{P(A, B, C=c, D, X=x)}{P(C = c \mid X = x)}$. This means the entropic inequalities $I(A(c) : B(c)) \le H(D(c))$ on this counterfactual distribution (one for each level of $C$) imply inequality constraints on the observed data distribution as well. These inequality constraints will be obtained in Figs.~\ref{fig:appendix2A} and \ref{fig:appendix2C}, but not in Fig.~\ref{fig:appendix2B}.
    
    Moreover, these inequality constraints can be shown to be nontrivial. Since Figs.~\ref{fig:appendix2A} and \ref{fig:appendix2B} share the same \(d\)-separation and \(e\)-separation relations it follows that any distributions compatible with Fig.~\ref{fig:appendix2B} cannot be witnessed as incompatible with Fig.~\ref{fig:appendix2A} using non-nested entropic equalities or inequalities. Consider the following structural equation model for Fig.~\ref{fig:appendix2B}: Let $U_1$, $U_2$ and $U_3$ be binary and uniformly distributed, and let ${X=U_2}$, ${A=U_2\oplus\epsilon_A}$, ${C=X\oplus U_3}$, ${B=C\oplus U_3\oplus\epsilon_B}$, and ${D=\epsilon_D}$ where \enquote{$\oplus$} indicates addition mod 2 and where $\epsilon_k$ is a random variable very heavily biased towards zero for $k\in \{A,B,D\}$. This establishes that $C$ and $X$ are uniformly distributed and statistically independent from each-other, and hence that $P(A,B)=P(A(c{=}0),B(c{=}0))$. This construction also gives ${A\oplus B} = {U_2\oplus \epsilon_A\oplus C\oplus U_3\oplus\epsilon_B}= {U_2 \oplus \epsilon_A\oplus X\oplus\epsilon_B}={\epsilon_A\oplus\epsilon_B}$ and hence ${A\approx B}$. This yields ${I(A(c{=}0) : B(c{=}0)) \approx H(A)\approx 1}$ whereas ${H(D(c{=}0))=H(D)\approx 0}$, strongly violating the entropic inequality ${I(A(c{=}0) : B(c{=}0)) \le H(D(c{=}0))}$ which applies only to Fig.~\ref{fig:appendix2A}.

\clearpage
\section{Proofs}
\label{app:proofs}

\textbf{Proof of Theorem~\ref{thm:entropic}}\par\nopagebreak

Let $\GS$ represent the graph in which every node in $\D$ is split, and $P^*$ denote
the distribution over variables in $\GS$. We follow the convention established above whereby for each node $D$ in $\D$, a new node $D^\#$ is added to the graph, $D^\#$ is made a parent of all children of $D$, and all edges outgoing from $D$ are removed.
For notational convenience, we let
$P_{\ds}(\cdot \mid \cdot) = P^*(\cdot \mid \cdot, \DS {=} \ds)$, and $I_{\ds}$
and $H_{\ds}$ be the mutual information and entropy in this counterfactual distribution. Recall
that by Theorem~\ref{thm:evans}, if \esep, then: 
\begin{compactenum}[4.i.]
\item\label{cond:zeroMI} ${I_{\ds}(\A : \B \mid \C {=} \c) = 0}$, and

\item\label{cond:consistency} ${{P(\A, \B, \D {=} \ds \mid \C {=}\c)}=
{P_{\ds}(\A, \B, \D {=} \ds \mid \C{=}\c)}}$. 
\end{compactenum}

From the latter condition \mbox{(4.\ref{cond:consistency}.)} we readily have that ${{H_{\ds}(\cdot \mid \cdot,\D{=}\ds)} = {H(\cdot \mid \cdot,\D{=}\ds)}}$. This means that for \emph{any} node set $\Z$, and for any node set $\X$ among the nondescendants of $\D$ in $\G$, we find that
\begin{align}
H(\Z\mid\X{=}\x,\D)&={\textstyle\sum_{\ds}} P(\D{=}\ds\mid \X{=}\x) H(\Z\mid\X{=}\x,\D{=}\ds) \nonumber\\
&={\textstyle\sum_{\ds}} P(\D{=}\ds\mid \X{=}\x) H_{\ds}(\Z\mid\X{=}\x,\D{=}\ds)\nonumber\\
&={\textstyle\sum_{\ds}} P_{\ds}(\D{=}\ds\mid \X{=}\x) H_{\ds}(\Z\mid\X{=}\x,\D{=}\ds)\label{eq:origcondD}
\end{align}
where the final equality follows from the fact that $P$ and $P^*$ agree when referring to nondescendants of $\DS$ in $\GS$, and that $\D$ is included among the nondescendants of $\DS$ in $\GS$.

If the counterfactual entropies only involve variables which are nondescendants of $\DS$ in $\GS$, then they coincide with the observable entropies even without such a weighted summation. This is because by construction $\DS$ has no parents, and therefore its nondescendants are marginally independent of $\DS$, and cannot be made dependent on it by conditioning on other nondescendants. 
When $\Y$ are among the nondescendants of $\DS$ in $\GS$ and further $\X$ are among among the nondescendants of $\D$ in $\G$, we find that
\begin{align}%
H(\Y\mid \X{=}\x)&=%
H_{\ds}(\Y\mid\X{=}\x),\nonumber\\
\shortintertext{which for later convenience we express as}
&=\left({\textstyle\sum_{\ds}} P_{\ds}(\D{=}\ds\mid \X{=}\x)=1\right)\times H_{\ds}(\Y\mid\X{=}\x).
\label{eq:strongnondescendants}
\end{align}
For $\X$ among the nondescendants of $\D$ in $\G$ we further introduce the notation\footnote{Note that the quantity defined in Eq.~\eqref{eq:surprisal} is a standard unit in information theory, known as ``surprisal" or ``self-information".}
\begin{subequations}\label{eq:justD}\begin{align}
h(\D=\ds\mid\X{=}\x)&=h_{\ds}(\D=\ds\mid\X{=}\x)\coloneqq -\log(P_{\ds}(\D{=}\ds\mid \X{=}\x))\label{eq:surprisal}
\shortintertext{such that}\label{eq:chaincondD}
H(\D\mid\X{=}\x)=H_{\ds}(\D\mid\X{=}\x)&={\textstyle\sum_{\ds}} P_{\ds}(\D{=}\ds\mid \X{=}\x) h_{\ds}(\D{=}\ds\mid\X{=}\x)
\end{align}\end{subequations}

To find constraints on the observed data distribution that take advantage of the properties of $P^*$, we can chose a particular node set $\X$ among the nondescendants of $\D$ in $\G$ and then identify entropic inequalities such that all the terms look like either 
\begin{align*}
&\text{Per~\eqref{eq:origcondD}:}&& H(\Z\mid\D,\X{=}\x) &=&{\textstyle\sum_{\ds}} P_{\ds}(\D{=}\ds\mid \X{=}\x) H_{\ds}(\Z\mid\D{=}\ds,\X{=}\x) \text{ for any }\Z,\\
&\text{Per~\eqref{eq:strongnondescendants}:}&&H(\Y\mid\X{=}\x) &=&{\textstyle\sum_{\ds}} P_{\ds}(\D{=}\ds\mid \X{=}\x) H_{\ds}(\Y\mid\X{=}\x)
\\\nonumber &&&&&\text{ for $\Y$ among nondescendants of $\DS$ in $\GS$ and $\Y\cap\D=\emptyset$},\\
&\text{Per~\eqref{eq:justD}:}&& H(\D\mid \X{=}\x) &=&{\textstyle\sum_{\ds}} P_{\ds}(\D{=}\ds\mid \X{=}\x)h_{\ds}(\D{=}\ds\mid\X{=}\x).
\end{align*}
Formally, the task of extracting all the implications of a set of linear inequalities on the subset of terms which have suitable form is an instance of polytope projection, and may be solved be means of Fourier-Motzkin elimination or related algorithms~\citep{Glle2018}.\footnote{As the example proofs demonstrate, the equality constraint(s) coming from \(e\)-separation are not in the desired final format. We identify all the undesirable terms, and then cancel them out by combining with subadditivity inequalities, monotonicity inequalities, as well as potentially any zero-mutual-information equalities associated with \(d\)-separation relations among the visible variables in the original graph $\G$ in order to obtain their testable implications.}

\begin{minipage}{\linewidth} 
To prove the first part of Theorem \ref{thm:entropic} in this way, suppose $\C$ are among the nondescendants of $\D$ in $\G$. Introducing the shorthand $\c$ to denote $\C=\c$ we could derive an inequality of suitable form by summing the four inequalities
\end{minipage}
\begin{subequations} \label{eqs:firstproof}
\begin{align}
& 0\leq I(\A:\D\mid \c) = H(\A\mid \c)-H(\A\mid \c,\D) \nonumber\\\label{eq:subaddA}
&\therefore\quad 0\leq -H(\A\mid \c,\D) + {\textstyle\sum_{\ds}}P_{\ds}(\D{=}\ds) H_{\ds}(\A\mid \c),\\[6pt]
& 0\leq I(\B:\D\mid \c) = H(\B\mid \c)-H(\B\mid \c,\D) \nonumber\\\label{eq:subaddB}
&\therefore\quad 0\leq  -H(\B\mid \c,\D) + {\textstyle\sum_{\ds}}P_{\ds}(\D{=}\ds) H_{\ds}(\B\mid \c),\\[6pt]
& 0\leq H(\D\mid \A,\B,\c) = H(\D\mid \c)+ H(\A,\B\mid \c,\D) - H(\A,\B\mid \c) \nonumber\\\label{eq:Dmonotonicity}
&\therefore\quad 0\leq  H(\D\mid \c)+ H(\A,\B\mid \c,\D) - {\textstyle\sum_{\ds}}P_{\ds}(\D{=}\ds) H_{\ds}(\A,\B\mid \c),\\[6pt]
& 0 = I_{\ds}(\A:\B\mid \c) = -{\textstyle\sum_{\ds}}P_{\ds}(\D{=}\ds) I_{\ds}(\A:\B\mid \c) \nonumber\\
&\therefore\quad 0\leq  {\textstyle\sum_{\ds}}P_{\ds}(\D{=}\ds) \left(H_{\ds}(\A,\B\mid \c) - H_{\ds}(\A\mid \c) - H_{\ds}(\B\mid \c)\right).\label{eq:zeroCIinstarworld}
\end{align}
\end{subequations}
Inequalities~\eqref{eq:subaddA} and~\eqref{eq:subaddB} follow from the nonnegativity of conditional mutual information. Subadditivity holds for both discrete and continuously valued variables~\cite{subadditivity}. Inequality~\eqref{eq:Dmonotonicity} follows from monotonicity (the fact that all conditional entropies are nonnegative) and the chain rule. Conditional entropy is only guaranteed to be nonnegative for discrete variables, and this is why  Theorem~\ref{thm:entropic} demands that $\D$ be discrete. All but the final inequality implicitly make use of Equation~\eqref{eq:origcondD}. Inequality~\eqref{eq:zeroCIinstarworld}, by contrast, is a consequence of ${I_{\ds}(\A : \B \mid \c) = 0}$
per Theorem~\ref{thm:evans}; see condition \mbox{(4.\ref{cond:zeroMI}.)} above.
Summing all four inequalities~\eqref{eqs:firstproof} leads to the derived inequality
\begin{align}
0\leq H(\D\mid \c) + H(\A,\B\mid \c, \D)-H(\A\mid \c,\D)-H(\B\mid \c,\D),
 \quad\text{i.e.,}\quad &I(\A:\B \mid \c, \D)\leq H(\D\mid \c).
\end{align}

Now consider the case where we are further promised that $\A$ are nondescendants of $\D$ in $\G$ and hence nondescendants of $\DS$ in $\GS$. This means that in addition to the above results we also have that $H_{\ds}(\A\mid \c)=H(\A\mid\c)$ per Equation~\eqref{eq:strongnondescendants}.
We proceed as before, but instead of summing all four of the \eqref{eqs:firstproof} inequalities we only take the sum of the latter three. This then yields
\begin{align}
0\leq H(\D\mid \c) + H(\A,\B\mid \c, \D)-H(\A\mid \c)-H(\B\mid \c,\D) %
\quad\text{i.e.,}\quad 
I(\A:\B,\D \mid \c) \leq H(\D\mid \c).
\end{align}

In both cases, the constraint is maintained after taking the expectation of both
sides with respect to $\C$. Because each term in the expectation will satisfy
the inequality, so will the sum. That is, in the first case, Eq.~\eqref{eq:firststrong} implies Eq.~\eqref{eq:firstweak}, and in the second case, Eq.~\eqref{eq:boundforMME} implies Eq.~\eqref{eq:secondweak}. \qed

\textbf{Beyond Theorem~\ref{thm:entropic}}\par\nopagebreak
Going beyond the result in Theorem~\ref{thm:entropic}, even when $\C$ are not entirely among the nondescendants of $\D$ in $\G$ we can nevertheless obtain the nontrivial entropic inequality ${I(\A:\B\mid \C,\D)\leq H(\D)}$, by noting that
\begin{align}
0\leq& I(\A:\D\mid \C)+I(\B:\D\mid \C)+H(\D\mid \A,\B,\C)+I(\C:\D)-{\textstyle\sum_{\ds}}P_{\ds}(\D{=}\ds) I_{\ds}(\A:\B\mid \C).
\end{align}
One notes that all of the terms introduced by $-{\textstyle\sum_{\ds}}P_{\ds}(\D{=}\ds) I_{\ds}(\A:\B\mid \C)$ are cancelled by judiciously applying Equation~\eqref{eq:origcondD} to the other terms, leaving 
\begin{align}
0\leq H(\D) + H(\A,\B,\C\mid \D)-H(\A,\C\mid \D)-H(\B,\C\mid \D)+H(\C\mid \D) = H(\D)-I(\A:\B\mid \C,\D).
\end{align}

Note that this proof technique can be adapted to derive stronger entropic inequalities for graphs which exhibit multiple different \(e\)-separation relations involving the same $\D$ set. If ${(\A_1\perp_e \B_1\mid\C_1 \text{ upon } \lnot \D)}$ and ${(\A_2\perp_e \B_2\mid\C_2 \text{ upon } \lnot \D)}$ and so forth, then Theorem~\ref{thm:evans} still demands the existence of a \emph{single} $P_{\ds}$ whose various margins must now satisfy \emph{multiple distinct} zero conditional mutual informational equalities. We can accommodate multiple entropic equality constraints on  $P_{\ds}$ just as easily as we can accommodate a single equality constraint: The translation between constraints on $P_{\ds}$ and $P$ will continue to be governed by conditions~\eqref{eq:origcondD},~\eqref{eq:strongnondescendants},~and~\eqref{eq:chaincondD}.

\begin{minipage}{\linewidth}
\textbf{Proof of Proposition~\ref{prop:d-entails-e}}\par\nopagebreak
If conditioning on some variables $\D$ is sufficient to close a path, then that
path must go through $\D$, and therefore deletion of $\D$ eliminates the path.
By construction, the deletion operation can never open a path, unlike the
conditioning operation. If ${(\A \perp_d \B \mid \C, \D)}$, then all paths
from $\A$ to $\B$ go through $\C$ or $\D$, or through colliders that are not
in $\{\C,  \D\}$, nor have any descendants therein. It follows that \esep, as
after deletion of $\D$ all paths through $\C$ remain blocked through
conditioning, all paths through $\D$ are eliminated, and all other paths
remain blocked by colliders.
\end{minipage}

\begin{minipage}{\linewidth}
\textbf{Proof of Proposition~\ref{prop:entropic-dsep}}\par\nopagebreak
Firstly, we note that the data processing inequality
\begin{align}
I({\A : \Ub \mid \C{=}\c}) &\ge I(\A : \B \mid \C{=}\c)\quad\text{whenever}\quad{\A \perp \B \mid \{\C, \Ub\}}
\end{align}
holds for continuous variables, since it is merely a consequence of the fact that the chain rule can by applied to %
 $I({\A : \B,\Ub \mid \c})$ in either of two ways, i.e. 
 \begin{align*}
  &I({\A : \B,\Ub \mid \c})= I({\A : \B \mid \c})+I({\A : \Ub \mid \B,\c}),
 \\\text{and also that}\quad
 &I({\A : \B,\Ub \mid \c})=I({\A : \Ub \mid \c})+I({\A : \B \mid \Ub,\c}).
 \end{align*}
 The data processing inequality, then, follows from the nonnegativity of $I({\A : \Ub \mid \B,\c})$ and from $I({\A : \B \mid \Ub,\c}){=}0$ being an implication of ${\A \perp \B \mid \{\C, \Ub\}}$. 
 Recognizing that $I({\A : \Ub \mid \c})={H(\Ub\mid\c)}-\left({H(\A,\Ub\mid\c)}-{H(\A\mid \c)}\right)$, and that by weak monotonicity ${H(\A,\Ub\mid\c)}-{H(\A\mid \c)}\geq 0$, we then have that 
 \begin{align}
 {H(\A\mid \C{=}\c)}\geq I({\A : \Ub \mid \C{=}\c}).
 \end{align}

To obtain Corollary~\ref{cor:latent-cardinality} simply note that the constraint is maintained after taking the expectation of both
sides with respect to $\C$: because each term in the expectation will satisfy
the inequality whenever the cardinality of $\Ub$ is finite, so will the sum.
\end{minipage}

\textbf{Proof of Corollary~\ref{cor:mme-bound}}\par\nopagebreak
Per Definition~\ref{def:MME}, $\MME_{X\to Y}$ is defined as the smallest entropy $H(W)$ over all structural equations models over $\G^\prime_{X\to W\to Y}$ in which $W$ has finite cardinality and which reproduce the observed data distribution over $\{\A,\B,\C,\D\}$. Let $P'(\A,\B,\C,\D,W)$ be the distribution over $\{\A,\B,\C,\D,W\}$ associated with some structural equation model which minimizes $H(W)$. Since the premises of Corollary~\ref{cor:mme-bound} stipulate that $\G^{\prime}_{X \to W \to Y}$  exhibits the \(e\)-separation relation ${(\A\perp_e \B\mid\C \text{ upon } \lnot \{\D,W\})}$, and that no element of $\{\A,\C\}$ is a descendant of any in $\{\D,W\}$, it follows from Equation~\eqref{eq:secondweak} in our main theorem that $P'$ must satisfy $I(\A:\B,W\mid\C,\D)\leq H(\D,W\mid \C)$. Indeed, one can confirm the following sequence of inequalities which $P'$ must satisfy:

\begin{subequations}\begin{align}
  I(\A:\B \mid\C,\D)
  &\leq I(\A:\B,W\mid\C,\D)\label{eq:pretheoremapplied}\\
  &\leq H(\D,W\mid \C)\label{eq:posttheoremapplied}\\
  &\leq H(\D\mid \C)+ H(W\mid \C)\label{eq:MMEproofspecificC}\\
  &\leq H(\D\mid \C) + H(W)\label{eq:MMEproofend}\\
  &= H(\D\mid \C) + \MME_{X \to Y}\nonumber
  \end{align}\end{subequations}
 where all the steps above are consequences of subadditivity except for the step from Equation~\eqref{eq:pretheoremapplied} to Equation~\eqref{eq:posttheoremapplied}, which is just the application of Equation~\eqref{eq:secondweak} as noted above. %

\nopagebreak
Note that we have elected to define $\MME\coloneqq \min_{\text{SEM }\G^\prime}\;H(W)$, i.e. Definition~\ref{def:MME} minimizes $H(W)$ over all structural equations models over $\G^\prime_{X\to W\to Y}$. One may also consider distinct but related entropic measures of causal influence to $Y$ from $Y$, such as 
\begin{align}
\MME'\coloneqq \max\nolimits_{\Z,\z}\; \min\nolimits_{\text{SEM }\G^\prime}\; H(W\mid \textbf{do }\Z{=}\z), 
\end{align}
which involves recomputing (or at least lower bounding) $\MME$ under different external intervention choices ($\Z$ being any subset of the variables in $G$ excluding $X$ and $Y$, and $\z$ being any value tuple for $\Z$) and retaining the largest of all the lower bounds. Plainly $\MME'\geq \MME$. 
However, we can exploit Equation~\eqref{eq:boundforMME} to obtain a slightly stronger lower bound on $\MME'$ than just  ${\MME_{X \to Y}\geq I(\A : \B \mid \C,\D)-H(\D\mid \C)}$ when the premises of Corollary~\ref{cor:mme-bound} hold, namely
\begin{align}
\MME' \geq \max\nolimits_{\c} \;I(\A : \B \mid \D, \textbf{ do }\C{=}\c)-H(\D\mid \textbf{ do }\C{=}\c).
\end{align}

\begin{samepage}
\section{Relation between Common Entropy and \MME} 

The \MME bears some resemblance to a concept called \emph{common entropy} \citep{commonentropy}, which is defined for a distribution $P(X, Y)$ as the smallest possible entropy of an unobserved variable $W$ such that $X \perp Y \mid W$. Unlike the \MME, the common entropy is a function only of the probability distribution $P(X, Y)$, and not of the graph $\mathcal G$. Any $W$ that renders $X$ and $Y$ conditionally independent must also fully mediate the effect of $X$ on $Y$, which at first glance might be taken to mean that the common entropy is an upper bound on the \MME, because it implies that the \MME can search over a larger set of distributions to obtain a low-entropy mediator. Indeed in the simple ${X \to Y}$ model, it is the case that $\MME_{X \to Y}$ is bounded from above by the common entropy between $X$ and $Y$ for precisely this reason. 

\nopagebreak
However, the common entropy is not an upper bound on the \MME in general. To see this, consider the graph presented in Fig. \ref{fig:node-splitting}(c). This model contains distributions in which $A$ and $B$ are highly correlated, but $D$ and $B$ are entirely uncorrelated. For such distributions, the common entropy of $B$ and $D$ would be $0$, as they are already marginally independent. However, by Corollary \ref{cor:mme-bound}, the \MME would be bounded from below by $I(A:B)$, which can be larger than $0$. The intuition for this phenomenon is that if the edge $D \rightarrow B$ were missing, $A$ and $B$ would be marginally independent, so a high mutual information between them is evidence for the causal significance of the edge.
\end{samepage}

\clearpage
\section{An inverse of Theorem~\lowercase{\ref{thm:entropic}}}\label{app:inequivalence}

The goal of this appendix is to establish that
\begin{prop}\label{prop:canviolate}
If a graph $\G$ has the feature $(\A\not\perp_e \B\mid\C \text{ upon } \lnot \D)$, then there exists a distribution in the marginal model of $\G$ with discrete $\D$ such that
${(A:B\mid\C,\D)=I(A:B\mid \C)\gneq H(\D)}$.
\end{prop}

We begin by simply noting that

\begin{lemma}\label{lem:ignoreD}
The marginal model of \emph{any} graph $\G$ whose nodes include $\{\A,\B,\C,\D\}$ contains all conditional distributions $P(\A,\B\mid\C,\D)$ wherein
\begin{compactenum}
\item $P(\A,\B\mid\C,\D)=P(\A,\B\mid\C)$, and
\item $P(\A,\B\mid\C)$ is within the marginal model of the graph $\G'$ defined by removing outgoing edges from $\D$ in $\G$.
\end{compactenum}
\end{lemma}
The sorts of $P(\A,\B\mid\C,\D)$ described in Lemma~\ref{lem:ignoreD} arise by considering causal models wherein every child of $\D$ always ignores the values of $\D$, treating $\D$ as if it has no descendants.
We next invoke the completeness of $d$-separation. That is,
\begin{lemma}\label{lem:dsep_weak_alternative}
If  $(\A\not\perp_d \B\mid\C)$ in $\G$, then there exists a distribution in the marginal model of $\G$ for which ${I(\A:\B\mid\C)\gneq 0}$.
\end{lemma}
We note that Lemma~\ref{lem:dsep_weak_alternative} follows from 
\begin{lemma}\label{lem:dsep_weak}
If $(A\not\perp_d B\mid\C)$ in $\G$ for singleton nodes $A$ and $B$, then there exists a distribution in the marginal model of $\G$ for which $\exists_\c\text{ s.t. }{I(A:B\mid\C{=}\c)\gneq 0}$.
\end{lemma}
After all, ${I(\A:\B\mid\C)=0}$ if and only if ${I(A:B\mid\C)=0}$ for all singleton nodes $A\in\A$ and $B\in\B$. Moreover, ${I(A:B\mid\C)=0}$ if and only if ${I(A:B\mid\C{=}\c)=0}$ for all $\c$ having positive support. We believe that Lemma~\ref{lem:dsep_weak}
explicitly follows from Theorem 3 in Ref.~\cite{dsepcomplete}, but we provide an explicit proof of it below for completeness.

By combining Lemmas~\ref{lem:ignoreD} and~\ref{lem:dsep_weak_alternative} we obtain Proposition~\ref{prop:canviolate}. This follows by noting that whenever a graph $\G$ has the feature $(\A\not\perp_e \B\mid\C \text{ upon } \lnot \D)$, then \emph{by definition} the graph $\G'$ defined by removing outgoing edges from $\D$ in $\G$ exhibits $(\A\not\perp_d \B\mid\C)$. To violate the basic inequality in Theorem~\ref{thm:entropic} we apply Lemma~\ref{lem:dsep_weak_alternative} while keeping $H(\D)\lneq I(\A:\B\mid\C)$.  We can make $H(\D)$ arbitrarily small by heavily biasing $P(\D)$ towards one value.

\bigskip
\hrule
\textbf{Proof of Lemma~\ref{lem:dsep_weak}}~\\
The following construction yields ${I(A:B\mid\C{=}\mathbf{1})=\log\left(2\right)}$ whenever $\G$ exhibits $(A\not\perp_d B\mid\C)$ for singleton nodes $A$ and $B$.

If $(A\not\perp_d B\mid\C)$ in $\G$ then there exists some \emph{path} in $\G$ with end nodes $A$ and $B$ such that all colliders in the path are elements of $\C$ and no element in $\C$ is present in the path \emph{except} as a collider.
We classify the nodes within the path into three distinct types:
\begin{compactdesc}
\item[Two Incoming Edges from the Path] These are the colliders in the path, elements of $\C$. We take each such node to act as a Kronecker delta function over its two in-path parents. That is, it will return the value $1$ iff the two in-path parents have coinciding values. That is,
\begin{align}
y(x_1,x_2)&=\begin{cases}
0, & \text{with unit probability iff }x_1\neq x_2,
\\1 & \text{with unit probability iff }x_1=x_2.
\end{cases}
\intertext{\item[One Incoming Edge from the Path] These are the mediaries in the path, as well as at least one (perhaps both) of the end nodes of the path. Let these variables act as an identity functions of its single in-path parent. That is,}
y(x) &= \begin{cases}0&\text{with unit probability iff }x=0,\\
1&\text{with unit probability iff }x=1.
\end{cases}.
\intertext{\item[Zero Incoming Edges from the Path] These are the bases of forks in the path, as well as potentially one of the end nodes of the path. Let these variable act as uniformly random variables with cardinality 2. That is,}
y()&=\begin{cases}0 & \text{with probability }\frac{1}{2},\\1 & \text{with probability }\frac{1}{2}. \end{cases}
\end{align}
\end{compactdesc}
This construction results in every non-collider being uniformly distributed over $\{0,1\}$ and always taking the same value as every other non-collider in the path upon postelecting all colliders in the path to take the value $1$. That is, this construction explicitly ensures that ${I(A:B\mid\C{=}\mathbf{1})=\log\left(2\right)}$.

\end{document}